\documentclass[11pt,a4paper]{article}
\usepackage[hyperref]{emnlp2020}
\usepackage{times}
\usepackage{latexsym}

\usepackage{natbib}
\usepackage[utf8x]{inputenc}
\usepackage{comment}
\usepackage{todonotes}
\usepackage[acronym]{glossaries}
\usepackage{subcaption}
\usepackage{tabularx}
\usepackage{mathtools}
\usepackage{makecell}

\usepackage{microtype}

\aclfinalcopy % Uncomment this line for the final submission
\def\aclpaperid{2077} %  Enter the acl Paper ID here

\newcolumntype{Y}{>{\centering\arraybackslash}X}
\newcolumntype{L}[1]{>{\raggedright\let\newline\\\arraybackslash\hspace{0pt}}m{#1}}
\newcolumntype{C}[1]{>{\centering\let\newline\\\arraybackslash\hspace{0pt}}m{#1}}
\newcolumntype{R}[1]{>{\raggedleft\let\newline\\\arraybackslash\hspace{0pt}}m{#1}}
\DeclareUnicodeCharacter{2212}{-}

\title{Textual Supervision for Visually Grounded Spoken Language Understanding}

\author{Bertrand Higy\\
  Cognitive Science and AI\\
  Tilburg University \\
  \texttt{b.j.r.higy@uvt.nl} \\\And
  Desmond Elliott\\
  Department of Computer Science\\
  University of Copenhagen \\
  \texttt{de@di.ku.dk} \\\AND
  Grzegorz Chrupała \\
  Cognitive Science and AI\\
  Tilburg University \\
  \texttt{g.chrupala@uvt.nl}}

\date{}

\begin{document}

% Definitions of acronyms
\glsdisablehyper
\newacronym{ASR}{ASR}{automatic speech recognition}
\newacronym{GRU}{GRU}{gated recurrent unit}
\newacronym{MFCC}{MFCC}{mel-frequency cepstrum coefficient}
\newacronym{NLU}{NLU}{natural language understanding}
\newacronym{SLT}{SLT}{spoken language translation}
\newacronym{SLU}{SLU}{spoken language understanding}
\newacronym{WER}{WER}{word error rate}

\maketitle
\begin{abstract}
Visually-grounded models of spoken language understanding extract semantic
information directly from speech, without relying on transcriptions. This is
useful for low-resource languages, where transcriptions can be
expensive or impossible to obtain. Recent work showed that these models can be
improved if transcriptions are available at training time. However, it is
not clear how an end-to-end approach compares to a traditional
pipeline-based approach when one has access to transcriptions. Comparing
different strategies, we find that the pipeline approach works
better when enough text is available. With low-resource languages in mind, we
also show that translations can be effectively used in place of transcriptions
but more data is needed to obtain similar results.
\end{abstract}

\section{Introduction}

Spoken language understanding promises to transform our interactions
with technology by allowing people to control electronic devices
through voice commands. However, mapping speech to meaning is far from trivial. The traditional
approach, which has proven its effectiveness, relies on text as an intermediate
representation. This so-called pipeline approach combines an \gls{ASR} system
and a \gls{NLU} component. While this allows us to take advantage of improvements
achieved in both fields, it requires transcribed speech, which is an expensive
resource.

\begin{figure}[t!]
    \centering
    \includegraphics[scale=0.52]{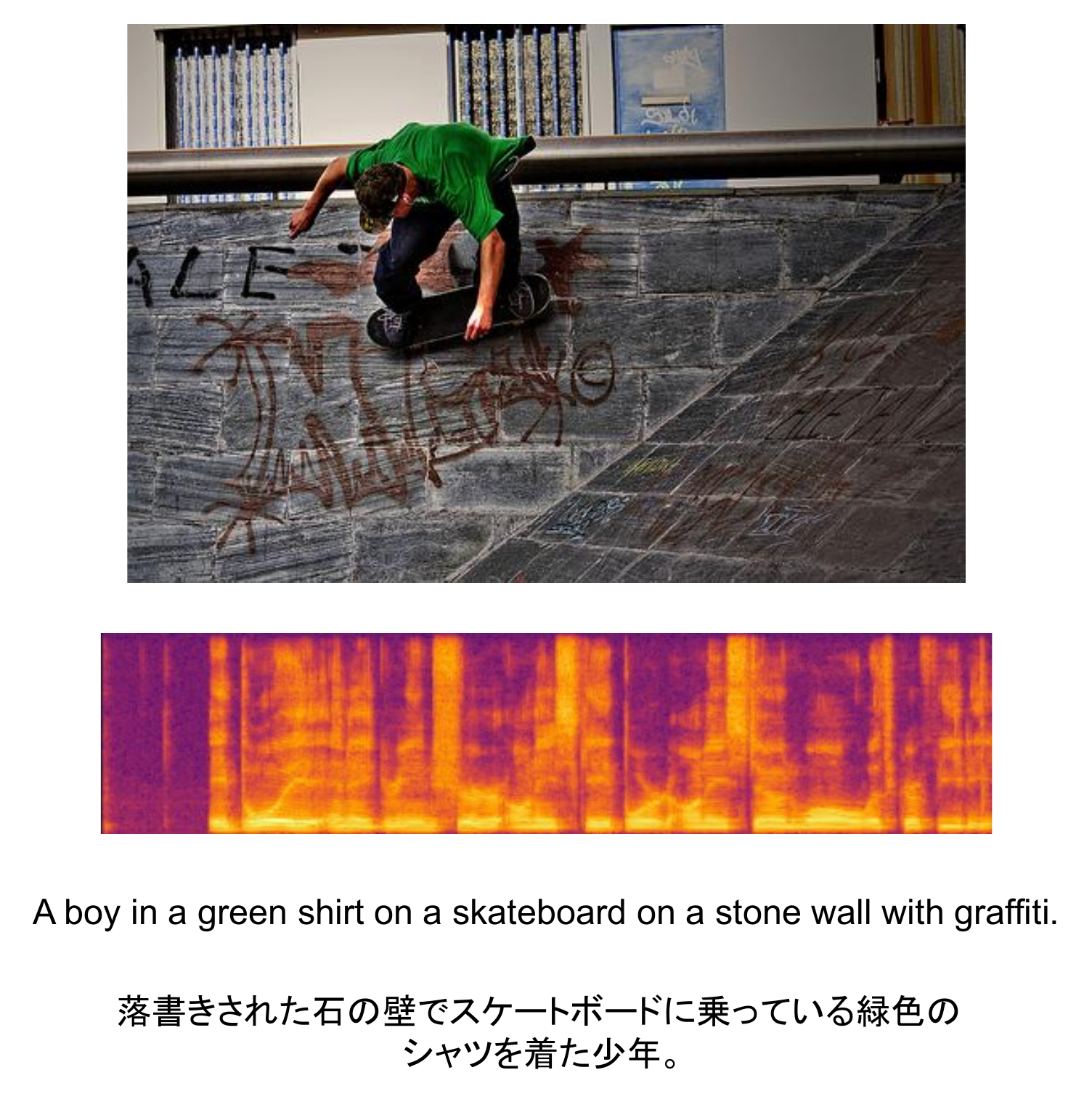}
    \caption{An image described by an English spoken caption (represented by
    its spectrogram), its transcription, and translation into
    Japanese. Visually-grounded models are usually trained to map the image and
    its spoken caption into a shared semantic space. }
    \label{fig:vgslu}
\end{figure}

Visually-grounded models of spoken language understanding
\citep{harwath_unsupervised_2016,chrupala_representations_2017} were recently
introduced to extract semantic information from speech directly, without
relying on textual information (see Figure \ref{fig:vgslu} for an
illustration). The advantages of these approaches are twofold: (i) expensive
transcriptions are not necessary to train the system, which is beneficial for
low-resource languages and (ii) trained in an end-to-end fashion, the whole
system is optimized for the final task, which has been shown to improve
performance in other applications.  While text is not necessary to train such
systems, recent work has shown that they can greatly benefit from textual
supervision if available \citep{chrupala_symbolic_2019,
pasad_contributions_2019}, generally using the multitask learning setup (MTL)
\citep{caruana_multitask_1997}.

However, the end-to-end MTL-based models in previous works have not been compared
against the more traditional pipeline approach that uses ASR as an intermediate
step. The pipeline approach could be a strong baseline as, intuitively, written
transcriptions are an accurate and concise representation of spoken language.
In this paper, we set out to determine the relative differences in performance
between end-to-end approaches and pipeline-based approaches. This study provides
insights from a pragmatic point of view, as well as having important
consequences for making further progress in understanding spoken language.

We also explore the question of the exact nature of the textual representations
to be used in the visually-grounded spoken language scenario. The text used in
previous work were {\it transcriptions}, which are a relatively faithful
representation of the form of the spoken utterance. Other possibilities
include, for example, subtitles, which tend to be less literal and abbreviated,
or translations, which express the meaning of the utterance in another
language. We focus on the case of translations due to the relevance of this
condition for low-resource languages: some languages without a standardized
writing system have many millions of speakers. One example is Hokkien, spoken
in Taiwan and Southeast China: it may be more practical to collect translations
of Hokkien into Mandarin than to get them transcribed. The question is whether
translations would be effective as a source of textual supervision in
visually-grounded learning.

In summary, our contributions are the following.
\begin{itemize}
    \item We compare different strategies for leveraging textual supervision in the
        context of visually-grounded models of spoken language understanding:
        we compare a pipeline approach, where speech is first
        converted to text, with two end-to-end MTL systems. We find that the
        pipeline approach tends to be more effective when enough textual data is available.
    \item We analyze how the amount of transcribed data affects performance, showing that end-to-end
        training is competitive only in very limited text conditions; however,
        textual supervision via transcribed data is marginally effective at
        this stage.
      \item We explore the possibility of replacing transcriptions with
        written translations.  In the case of translations, an
        end-to-end MTL approach outperforms the pipeline baselines; we
        also observe that more data is necessary with translations
        than with transcriptions, due to the more complex task faced
        by the system.
\end{itemize}

\section{Related work}

\subsection{Visually-grounded models of spoken language understanding}

Recent work has shown that semantic information can be extracted from speech in
a weakly supervised manner when matched visual information is available. This
approach is usually referred to as visually-grounded spoken language
understanding. While original work focused on single words
\citep{synnaeve_learning_2014, harwath_deep_2015}, the concept has quickly been
extended to process full sentences \citep{harwath_unsupervised_2016,
chrupala_representations_2017}. Applied on datasets of images with spoken
captions, this type of model is typically trained to perform a speech-image
retrieval task where an utterance can be used to retrieve an image which it is a
description of (or vice-versa). This is achieved through a triplet loss between
images and utterances (in both directions).

A similar approach is used by \citet{kamper_visually_2018} to perform semantic
keyword spotting. They include an image tagger in their model to provide tags
for each image in order to retrieve sentences that match a keyword semantically,
i.e.\ not only exact matches but also semantically related ones.

\subsection{Textual supervision}

A common thread in all of these studies is that the spoken sentences do not need
to be transcribed, which is useful due to the cost attached to textual labeling.
Subsequent work, however, showed that textual supervision, if available, can
substantially improve performance. \citet{chrupala_symbolic_2019} uses
transcriptions through multitask learning. He finds that adding a speech-text
matching task, where spoken captions have to be matched with corresponding
transcriptions, is particularly helpful. \citet{pasad_contributions_2019}
applied the same idea to semantic keyword spotting with similar results. They
also examine the effect of decreasing the size of the dataset.

\citet{hsu_transfer_2019} explore the use of visually grounded models to
improve \gls{ASR} through transfer learning from the semantic matching task; in contrast, we are interested in improving the performance of the grounded model
itself using textual supervision.

\subsection{Multitask versus pipeline}

Another area of related work is found in the spoken command understanding
literature. \citet{haghani_audio_2018} compare different architectures making
use of textual supervision, covering both pipeline and end-to-end approaches.
The models they explore include an ASR-based multitask system similar to the
present work.  For the pipeline system, they try  both independent and joint
training of the \gls{ASR} and \gls{NLU} components. Their conclusion is that an
intermediate textual representation is important to achieve good performance and
that jointly optimizing the different components improves predictions.
\citet{lugosch_speech_2019} propose a pretraining method for end-to-end spoken
command recognition that relies on the availability of transcribed data.
However, while this pretraining strategy brings improvement over a system
trained without transcripts, the absence of any other text-based baseline (such
as a pipeline system) prevents any conclusion on the advantage of the end-to-end
training when textual supervision is available.

\subsection{Multilingual data}

The idea of using multilingual data is not new in the literature: existing work
focuses on using the same modality for the two languages, either text or
speech. \citet{gella_image_2017} and \citet{kadar_lessons_2018} show that
textual descriptions of images in different languages in the Multi30K dataset
\cite{elliott2016multi30k} can be used in conjunction to improve the
performance of a visually-grounded model. \citet{harwath_vision_2018} focus on
speech, exploring how spoken captions in two languages can be used
simultaneously to improve performance in an English-Hindi parallel subset of
the Places-205 dataset \cite{zhou2014learning}. In contrast, our experiments
concern the setting where speech data from a low-resource language is used in
conjunction with corresponding translated written captions.

Directly mapping speech to textual translation, or
\gls{SLT}, has received increasing interest lately. Following recent trends in
\gls{ASR} and machine translation, end-to-end approaches in particular have
drawn much attention, showing competitive results against pipeline systems
\citep[e.g.][]{berard_listen_2016, weiss_sequence--sequence_2017}.

\section{Methodology}

The architecture and the training procedure used in this paper are inspired by
the improved version of the visually-grounded spoken language understanding
system of \citet{merkx_language_2019}. Appendix
\ref{sec:app_hyperparameters} provides details on the choice of
hyperparameters.

\subsection{Architectures}

\begin{figure}[t!]
    %\quad\quad
    \begin{subfigure}[t]{0.49\linewidth}
        \centering
        \includegraphics[scale=0.4]{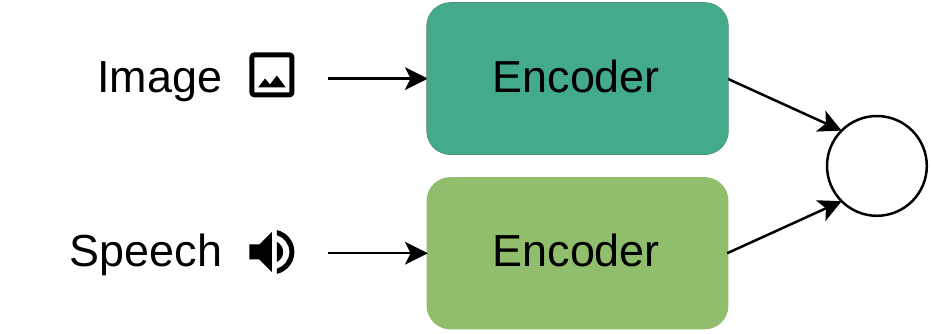}
        \caption{\emph{speech-image} model}
        \label{fig:diag_speech_image}
    \end{subfigure}
    %\quad
    %
    \begin{subfigure}[t]{0.49\linewidth}
        \centering
        \includegraphics[scale=0.4]{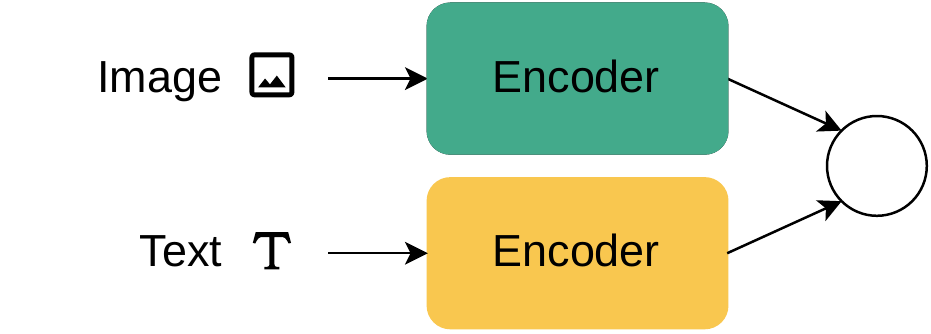}
        \caption{\emph{text-image} model}
        \label{fig:diag_text_image}
    \end{subfigure}
    \begin{subfigure}{\linewidth}
        \centering
        \vspace{10pt}
        \includegraphics[scale=0.5]{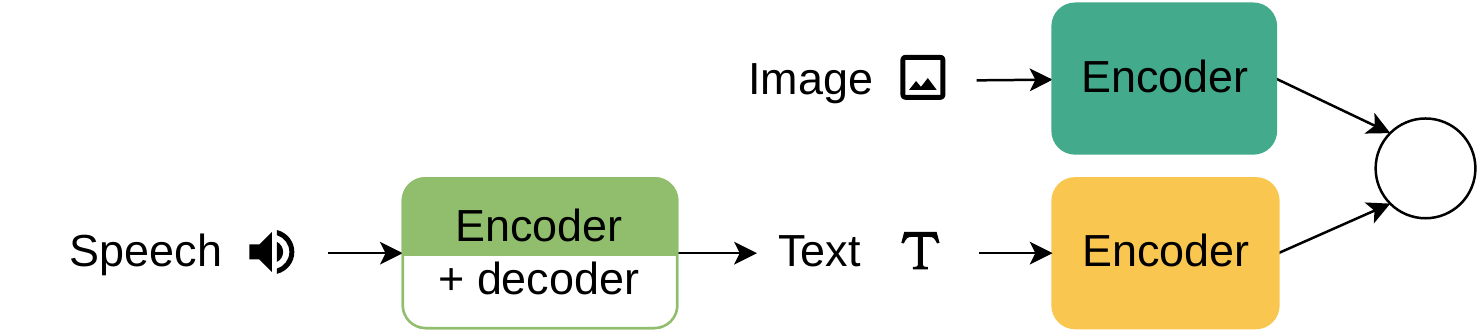}
        \caption{pipeline models (\emph{pipe-ind} and \emph{pipe-seq})}
        \label{fig:diag_pipeline}
    \end{subfigure}
    \hspace{\fill}
    \begin{subfigure}{\linewidth}
        \centering
        \vspace{10pt}
        \includegraphics[scale=0.5]{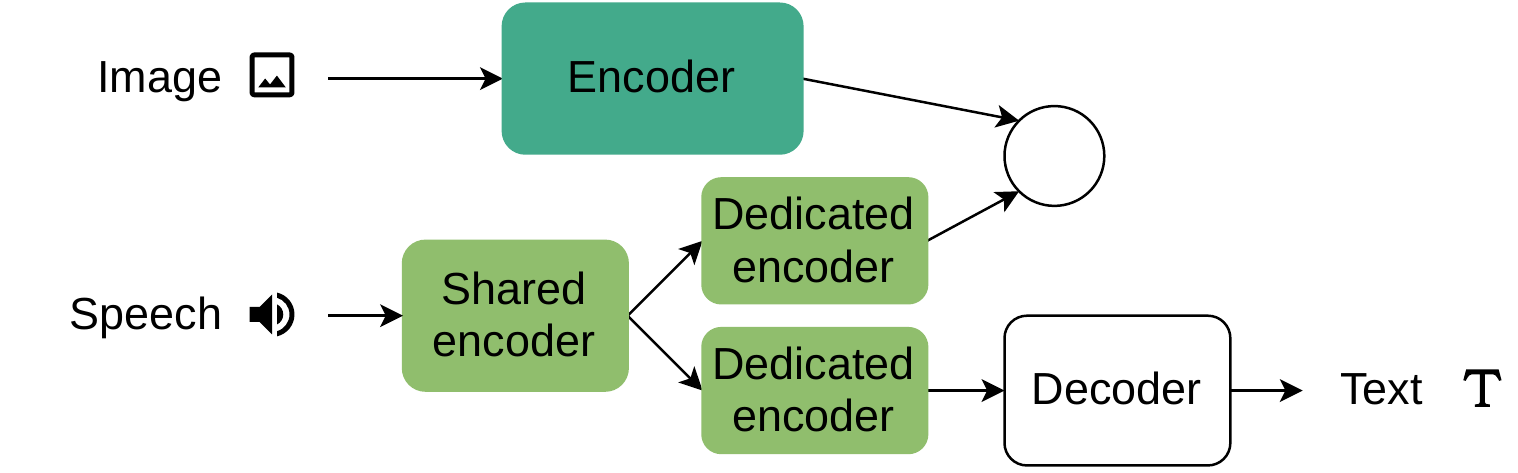}
        \caption{\emph{mtl-transcribe}/\emph{mtl-translate} model}
        \label{fig:diag_mtl_asr}
    \end{subfigure}
    \begin{subfigure}{\linewidth}
        \centering
        \vspace{10pt}
        \includegraphics[scale=0.5]{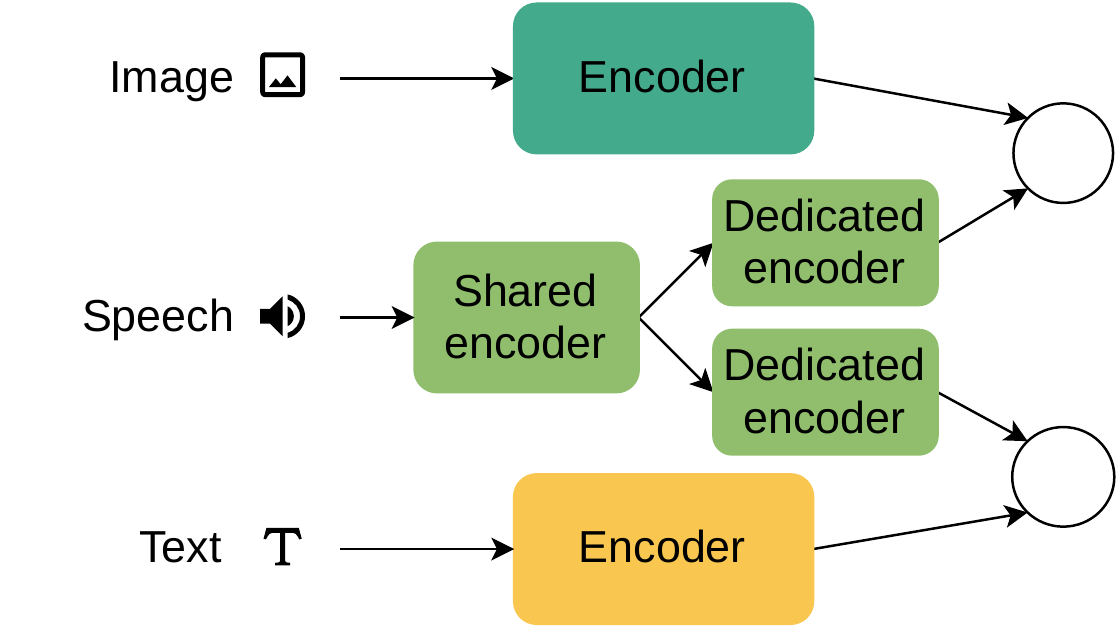}
        \caption{\emph{mtl-match} model}
        \label{fig:diag_mtl_st}
    \end{subfigure}
    \caption{Architecture of the different models.}
    \label{fig:architectures}
\end{figure}

We will compare six different models, based on five architectures
(summarized in Figure \ref{fig:architectures}).

Two models serve as reference, the original speech-image matching system
\citep{harwath_unsupervised_2016, chrupala_representations_2017} that does not
rely on text and a text-image model that works directly on text (and thus
requires text even at test time; similar to the \emph{Text-RHN} model from
\citet{chrupala_representations_2017} or the \emph{Char-GRU} model of
\citet{merkx_language_2019}).

We then have the four core models which  use text during training but
can also work with speech only at test time. Those are the four models we are
mainly interested in. They comprise: two pipeline models, which only differ in
their training procedure, and two multitask systems, using either a speech-text
retrieval task (similar to what is done in \citet{chrupala_symbolic_2019} and
\citet{kamper_visually_2018}) or \gls{ASR} as the secondary target.

\subsubsection{The speech-image baseline}
\label{sec:arch_si}

The speech-image baseline (\emph{speech-image}) is composed of two main
components: an image encoder and a speech encoder (see Figure
\ref{fig:diag_speech_image}).

\textbf{Image encoder.} The image encoder is composed of a single linear layer
projecting the image features (see Section \ref{sec:preproc}) into the shared
semantic embedding space (dimension 2048), followed by a normalization layer
($\ell^2$ norm).

\textbf{Speech encoder.} The speech encoder is applied on the \gls{MFCC}
features described in Section \ref{sec:preproc} and composed of a 1D
convolutional layer (kernel size 6, stride 2 and 64 output channels), followed
by bidirectional \glspl{GRU} \citet{cho_learning_2014} (4 layers, hidden
state of dimension 1024). A vectorial attention layer is then used to
convert the variable length input sequence to a fixed-size vector of dimension
2048. Finally, a normalization layer ($\ell^2$ norm) is applied.

\subsubsection{The text-image baseline}
\label{sec:arch_ti}

The text-image baseline (\emph{text-image}) measures the possible performance
if text is available at test time. It serves as a high estimate of what the
four core models could achieve.  Those models could theoretically perform
better than the \emph{text-image} baseline by taking advantage of information
available in speech and not in text. However, extracting the equivalent of the
textual representation from speech is not trivial so we expect them to perform
worse.

The \emph{text-image} model is comprised of an image encoder and a text encoder
(see Figure \ref{fig:diag_text_image}). The image encoder is identical to the
\emph{speech-image} model.

\textbf{Text encoder.} The text encoder is character-based and maps the input
characters to a 128-dimensional space through an embedding layer. The output is
then fed to a bidirectional \gls{GRU} (2 layers, hidden state of
dimension 1024). A vectorial attention mechanism followed by a normalization
layer ($\ell^2$ norm) summarizes the variable-length sequence into fixed-length
vector (dimension 2048).

\subsubsection{The pipeline models}

We trained two pipeline models (\emph{pipe-ind} and \emph{pipe-seq}) which only
differ in their training procedure (see Section \ref{sec:tr_pip}). The
architecture (summarized in Figure \ref{fig:diag_pipeline}) is basically
composed of an \gls{ASR} module which maps speech to text, followed by the
\emph{text-image} system we just described (our \gls{NLU} component). The same
architecture is used when training with Japanese captions, though the first
part is then referred to as the \gls{SLT} module.

\textbf{\Gls{ASR}/\gls{SLT} module.} The \gls{ASR}/\gls{SLT} module is an
attention-based encoder-decoder system which can itself be decomposed into two
sub-modules, an encoder and an attention-based decoder. The encoder is similar
to the speech encoder described above: it is composed of the same convolutional
layer followed by a bidirectional \gls{GRU} (5 layers, hidden state of
dimension 768) but lacks the attention and normalization layers. The
attention-based decoder uses a timestep-dependent attention mechanism
\cite{bahdanau_neural_2015} to summarize the encoded input sequence into
fixed-size context vectors (one per output token). The recurrent decoder
generates the output sequence one character at a time. At each time step, it
takes the current context vector and the previous character as input.  It is
composed of a unidirectional \gls{GRU} (1 layer, hidden state of dimension
768), a linear projection and the softmax activation layer.

\subsubsection{The \gls{ASR}/\gls{SLT}-based multitask model}

The \gls{ASR}/\gls{SLT}-based multitask model (\emph{mtl-transcribe} and
\emph{mtl-translate} respectively) combines the \emph{speech-image} model with
an \gls{ASR}/\gls{SLT} system (similar to the one used in the pipeline models).
To do so, the speech encoder of the \emph{speech-image} system and the encoder
of the \gls{ASR}/\gls{SLT} sytem are merged in a single speech encoder composed
of a shared network followed by two task-specific networks (see Figure
\ref{fig:diag_mtl_asr}). The image encoder and the attention-based decoder
being identical to the ones described previously, we will focus on the
partially-shared speech encoder. The multitask training procedure is
described in Section \ref{sec:tr_mtl}.

\textbf{The \emph{mtl-transcribe}/\emph{mtl-translate} speech encoder.} The
shared part of the speech encoder is composed of a convolutional layer (same
configuration as before) and a bidirectional \gls{GRU} (4 layers, hidden
state of dimension 768). The part dedicated to the secondary \gls{ASR}/\gls{SLT}
task is only composed of an additional bidirectional \gls{GRU} (1 layer,
hidden state of dimension 768). The part dedicated to the speech-image
retrieval task is composed of the same \gls{GRU} layer but also incorporates the
vectorial attention and normalization layers necessary to map the data into the
audio-visual semantic space.

\subsubsection{The text-as-input multitask model}

The other multitask model (\emph{mtl-match}) is based on
\citet{chrupala_symbolic_2019}. It combines the \emph{speech-image} baseline
with a speech-text retrieval task (see Figure \ref{fig:diag_mtl_st}). Images
and text are encoded by subnetworks identical to the image encoder described in
Section \ref{sec:arch_si} and the text encoder described in Section \ref{sec:arch_ti}
respectively.

\textbf{The \emph{mtl-match} speech encoder.} Similarly to the
\emph{mtl-transcribe}/\emph{mtl-translate} architecture, the speech encoder is
composed of a shared component and two dedicated parts. The shared encoder is
again composed of a convolutional layer (same configuration as before),
followed by a bidirectional \gls{GRU} (2 layers, hidden state of
dimension 1024). The part of the encoder dedicated to the \emph{speech-image}
task is composed of a \gls{GRU} (2 layers, hidden state of dimension
1024), followed by the vectorial attention mechanism and the normalization
layer ($\ell^2$ norm). Its counterpart for the speech-text task is only made
of the vectorial attention mechanism and the $\ell^2$ normalization layer.

\subsection{Training procedure}
\label{sec:tr_proc}

\subsubsection{Losses}

\textbf{Retrieval loss.} The main objective used to train our models is the
triplet loss used by \citet{harwath_deep_2015}. The goal is to map images and
captions in a shared embedding space where matched images and captions are
close to one another and mismatched elements are further apart. This is
achieved through optimization of following loss:

\setlength{\abovedisplayskip}{0pt} \setlength{\abovedisplayshortskip}{0pt}
\begin{align}
    \sum\limits_{\mathclap{\substack{(u,i) \\ (u',i') \ne (u,i)}}} \bigg(&\max(0, d(u,i) - d(u',i) + \alpha) + \nonumber \\
                                                                         &\max(0, d(u,i) - d(u,i') + \alpha)\bigg),
\end{align}
where $(u,i)$ and $(u',i')$ are each a pair of matching utterance and image
from the current batch, $d(\cdot,\cdot)$ is the cosine distance between encoded
utterance and image, and $\alpha$ is some margin (we use the value of $0.2$).

Similarly, a network can be trained to match spoken captions with corresponding
transcriptions/translations, replacing utterance and image pairs $(u,i)$ with
utterance and text pairs $(u,t)$.

\textbf{\Gls{ASR}/\gls{SLT} loss.} The \gls{ASR} and \gls{SLT} tasks are
optimized through the usual cross-entropy loss between the decoded sequence
(using greedy decoding) and the ground truth text.

\subsubsection{Training the pipeline systems}
\label{sec:tr_pip}

We use two strategies to train the pipeline systems:

\begin{itemize}
    \item The \textbf{independent} training procedure (\emph{pipe-ind} model),
        where each module (\gls{ASR}/\gls{SLT} and \gls{NLU}) is trained
        independently from the other. Here the text encoder is trained on ground-truth written captions or translations.
    \item The \textbf{sequential} training procedure (\emph{pipe-seq} model),
        where we first train the \gls{ASR}/\gls{SLT} module. Once done, we
        decode each spoken caption (with a beam search of width 10), and use
        the output to train the \gls{NLU} system. Doing so reduces the mismatch
        between training and testing conditions, which can affect the
        performance of the \gls{NLU} component.  This second procedure is thus
        expected to perform better than the independent training strategy.
\end{itemize}

\subsubsection{Multitask learning}
\label{sec:tr_mtl}

The \emph{mtl-transcribe}/\emph{mtl-translate} and \emph{mtl-match} strategies
make use of multitask learning through shared weights. To train the models, we
simply alternate between the two tasks, updating the parameters of each task in
turn.

\subsubsection{Optimization procedure}

The optimization procedure follows \citet{merkx_language_2019}. We use
the Adam optimizer \cite{kingma_adam:_2015} with a cyclic learning rate
\cite{smith_cyclical_2017} varying from $10^{−6}$ and $2\times10^{−4}$. All
networks are trained for 32 epochs, and unlike
\citeauthor{merkx_language_2019}, we do not use ensembling.

\subsection{Evaluation metrics}

Typical metrics for retrieval tasks are recall at $n$ (R@n with n $\in \{1, 5,
10\}$) or median rank (Medr). To compute these metrics, images and utterances are
compared based on the cosine distance between their embeddings
resulting in a ranked
list of images for each utterance, in order of increasing distance.
One can then compute the proportion of utterances for which the
paired image appears in the top $n$ images (R@n), or the median rank of the paired
image over all utterances. For brevity, we only report results with R@10 in the
core of the paper. The complete set of results is available in Appendix~\ref{sec:app_res}.

For \gls{ASR} and \gls{SLT}, we report \gls{WER} and BLEU score respectively,
using beam decoding in both cases (with a beam width of 10).

All results we report are the mean over three runs of the same experiment with
different random seeds.

\subsection{Experimental setup}

\subsubsection{Datasets}
\label{sec:methodo_datasets}

The visually grounded models presented in this paper require pairs of images
and spoken captions for training. For our experiments on textual supervision,
we additionally need the transcriptions corresponding to those spoken captions,
or alternatively a translated version of these transcripts. We obtain these
elements from a set of related datasets:
\begin{itemize}
    \item Flickr8K \citep{hodosh_framing_2013} offers 8,000 images of everyday
        situations gathered from the  website \href{www.flickr.com}{flickr.com} together with
        English written captions (5 per image) that were obtained through crowd
        sourcing.
    \item The Flickr Audio Caption Corpus \citep{harwath_deep_2015}, augments
        Flickr8K with spoken captions read aloud by crowd workers.
    \item F30kEnt-JP \citep{nakayama_visually-grounded_2020} provides Japanese
        translations of the captions (generated by humans). It covers the
        images and captions from Flickr30k \citep{young_image_2014}, a superset
        of Flickr8K, but only provides the translations of two captions per
        image.\footnote{Items from F30kEnt-JP and Flickr8K were matched based
        on exact matches between the English written captions in both
        datasets.  We also corrected for missing hyphens (e.g.\ "red
        haired" and "red-haired" are considered the same), leaving us with
    15,498 captions with Japanese transcription.}
\end{itemize}

In all experiments, we use English as the source language for our models. While
English is not a low-resource language, it is the only one for which we have
spoken captions. The low-resource setting with translations is thus a simulated setting.

To summarize, we have 8,000 images with 40,000 captions (five per image), in
both English written and spoken form (amounting to $\sim$34 hours of speech).
In addition, we have Japanese translations for two captions per image.

Validation and test sets are composed of 1,000 images from the original set
each (with corresponding captions), using the split introduced in
\citet{karpathy_deep_2015}. The training set is composed of the 6,000 remaining
images.

We additionally introduce a smaller version of the dataset available for
experiments with English transcriptions (later referred to as the {\it reduced}
English dataset), matching in size the one used for experiments with Japanese
translations (i.e.\ keeping only the sentences that have a translation, even
though we use transcriptions).

\subsubsection{Pre-processing}
\label{sec:preproc}

Image features are extracted from the pre-classification layer of a frozen
ResNet-152 model \cite{he_deep_2016} pretrained on ImageNet
\cite{deng_imagenet_2009}. We follow \citet{merkx_language_2019} and use features that are the result of taking
the mean feature vector over ten crops of each image.

The acoustic feature vectors are composed of 12 \glspl{MFCC} and log energy,
with first and second derivatives, resulting in 39-dimensional vectors. They
are computed over windows of 25 ms of speech, with 10 ms shift.

\begin{comment}
\subsubsection{Hyperparameters tuning}

We started from the hyperparameters reported in \citet{merkx_language_2019} for
the architecture and optimization procedure. The number of \gls{GRU} layers in
the text encoder and the speech encoder of the \gls{ASR}/\gls{SLT} We further
optimized the number of layers and units in the different models based on  f
the base speech and image encoders, as well as the training hyperparameters.
\end{comment}

\subsection{Repository}

The code necessary to replicate our experiments is available under Apache
License 2.0 at \href{https://github.com/bhigy/textual-supervision}{github.com/bhigy/textual-supervision}.

\section{Results}
\label{sec:results}

\subsection{Impact of the architecture}

We first look at the performance of the different models trained with the full
Flickr8K training set and English transcriptions.

As expected, using directly text as input, instead of speech,
makes the task much easier. This is exemplified by the difference between the \emph{speech-image} and \emph{text-image} models in Table \ref{tab:res_baselines}.

\begin{table*}[t]
    \centering
    \begin{tabular}{ lccc }
        \hline
        Model & Transcriptions (full) &  Transcriptions (reduced) & Translations \\
        \hline
        \emph{speech-image}  & 0.416 & 0.280 & 0.285 \\
        \emph{text-image}  & 0.702 & 0.653 & 0.626 \\
        \hline
    \end{tabular}
    \caption{Validation set R@10 of our two reference models when trained either with English transcriptions, a reduced version of the English dataset, or Japanese translations.}
    \label{tab:res_baselines}
\end{table*}

Table \ref{tab:res_core} reports the performance of
the four core models. We can notice that both pipeline and multitask architectures
can use the textual supervision to improve results over the text-less baseline,
though, pipeline approaches clearly have an advantage with a R@10 of 0.642.
Unlike what one could expect, training the pipeline system in a sequential way
does not bring any improvement over independent training of the modules (at
least with this amount of data).

\begin{table*}[t]
    \centering
    \begin{tabular}{ lccc }
        \hline
        Model & Transcriptions (full) &  Transcriptions (reduced) & Translations \\
        \hline
        \emph{pipe-ind} & \textbf{0.642} & 0.586 & 0.347 \\
        \emph{pipe-seq} & \textbf{0.642} & \textbf{0.598} & 0.345 \\
        \emph{mtl-transcribe}/\emph{mtl-translate} & 0.569 & 0.478 & \textbf{0.394} \\
        \emph{mtl-match}  & 0.451 & 0.356 & 0.337 \\
        \hline
    \end{tabular}
    \caption{Validation set R@10 of our four core models, when trained either
    with English transcriptions, a reduced version of the English dataset, or
    Japanese translations.}\label{tab:res_core}
\end{table*}

Comparing the two multitask approaches, we see that using ASR as a secondary
task (\emph{mtl-transcribe}) is much more effective than using text as another
input modality (\emph{mtl-match}).

Table \ref{tab:res_test} also reports performance of the best pipeline and the
best multitask model on the test set.  For completeness, Appendix
\ref{sec:app_res} reports the same results as Tables \ref{tab:res_baselines},
\ref{tab:res_core} and \ref{tab:res_test} with different metrics (R@1, R@5 and
Medr). Appendix \ref{sec:app_asr} reports the performance on the \gls{ASR}
task.

\begin{table}[t]
    \centering
    \tabcolsep=0.11cm
    \begin{tabular}{ lcc }
        \hline
        Model & Transcriptions & Translations \\
        \hline
        \emph{pipe-seq}  & 0.631 & 0.348 \\
        \makecell{\emph{mtl-transcribe}/\\\emph{mtl-translate}} & 0.559 & 0.392 \\
        \hline
    \end{tabular}
    \caption{Test set R@10 of the best pipeline (\emph{pipe-seq}) and the best
    multitask (\emph{mtl-transcribe}/\emph{mtl-translate}) models, when trained with all English transcriptions or all Japanese
    translations.}
    \label{tab:res_test}
\end{table}

\subsection{Using translations}
\label{sec:res_trl}

Tables \ref{tab:res_baselines} and \ref{tab:res_core} report the performance of
the same models when trained with Japanese transcriptions. The scores are
overall lower than results with English transcriptions, which can be explained
by two factors: (i) the size of the dataset which is only
$\sim$2/5\textsuperscript{th} of the original Flickr8K (as evidenced by the
lower score of the text-less \emph{speech-image} baseline) and (ii) the added
difficulty introduced by the translation over transcriptions. Indeed, to
translate speech, one first needs to recognize what is being said and then
translate to the other language: thus translation involves many complex
phenomena (e.g.\ reordering) which are missing from the transcription task.

While the four strategies presented in Table \ref{tab:res_core} improve over
the \emph{speech-image} baseline, their relative order differ from what is
reported with English text. This time, the \emph{mtl-translate} approach is the
one giving the best score with a R@10 of $0.394$, outperforming the pipeline
systems (both performing similarly well in this context).

The difference in relative order of the models is likely the result of the
degraded conditions (less data and harder task) impacting the translation task
more severely than the speech-image retrieval task. The pipeline approaches,
which rely directly on the output of the \gls{SLT} component, are affected more
strongly than the \emph{mtl-translate} system where \gls{SLT} is only a
secondary target. This is in line with the results reported in the next section
on downsampling the amount of textual data.

Table \ref{tab:res_test} reports performance of
the best pipeline and the best multitask model on the test set.  For
completeness, Appendix \ref{sec:app_res} reports the same results as Tables
\ref{tab:res_baselines}, \ref{tab:res_core} and \ref{tab:res_test} with
different metrics (R@1, R@5 and Medr). Appendix \ref{sec:app_asr} reports the
performance on the \gls{SLT} task.

\subsection{Disentangling dataset size and task factor}

In an attempt to disentangle the effects of the smaller dataset and the harder
task, we also report results on the reduced English dataset described in
Section \ref{sec:methodo_datasets} (Table \ref{tab:res_baselines} and
\ref{tab:res_core}, 3\textsuperscript{rd} column) . Looking first at Table
\ref{tab:res_core}, we can see that both factors do indeed play a role in the
drop in performance, though not to the same extent. Taking \emph{pipe-seq}
model as example, reducing the size of the dataset results in a 7\% drop in
R@10, while switching to translations further reduces accuracy by 42\%.

An unexpected result comes from the \emph{text-image} system (Table
\ref{tab:res_baselines}). Even though the model works on ground-truth text (no
translation involved), we still see a 4\% drop in R@10 between the reduced
English condition and Japanese. This suggests that the models trained with
Japanese translations are not only penalized by the translation task being
harder, but also that extracting meaning from Japanese text is more challenging
than from English (possibly due to a more complicated writing system).

\subsection{Downsampling experiments}

We now report on experiments that downsample the amount of
textual supervision available while keeping the amount of speech and images
fixed.

\begin{figure*}[t!]
    \begin{subfigure}[t]{0.5\linewidth}
        \includegraphics[width=\linewidth]{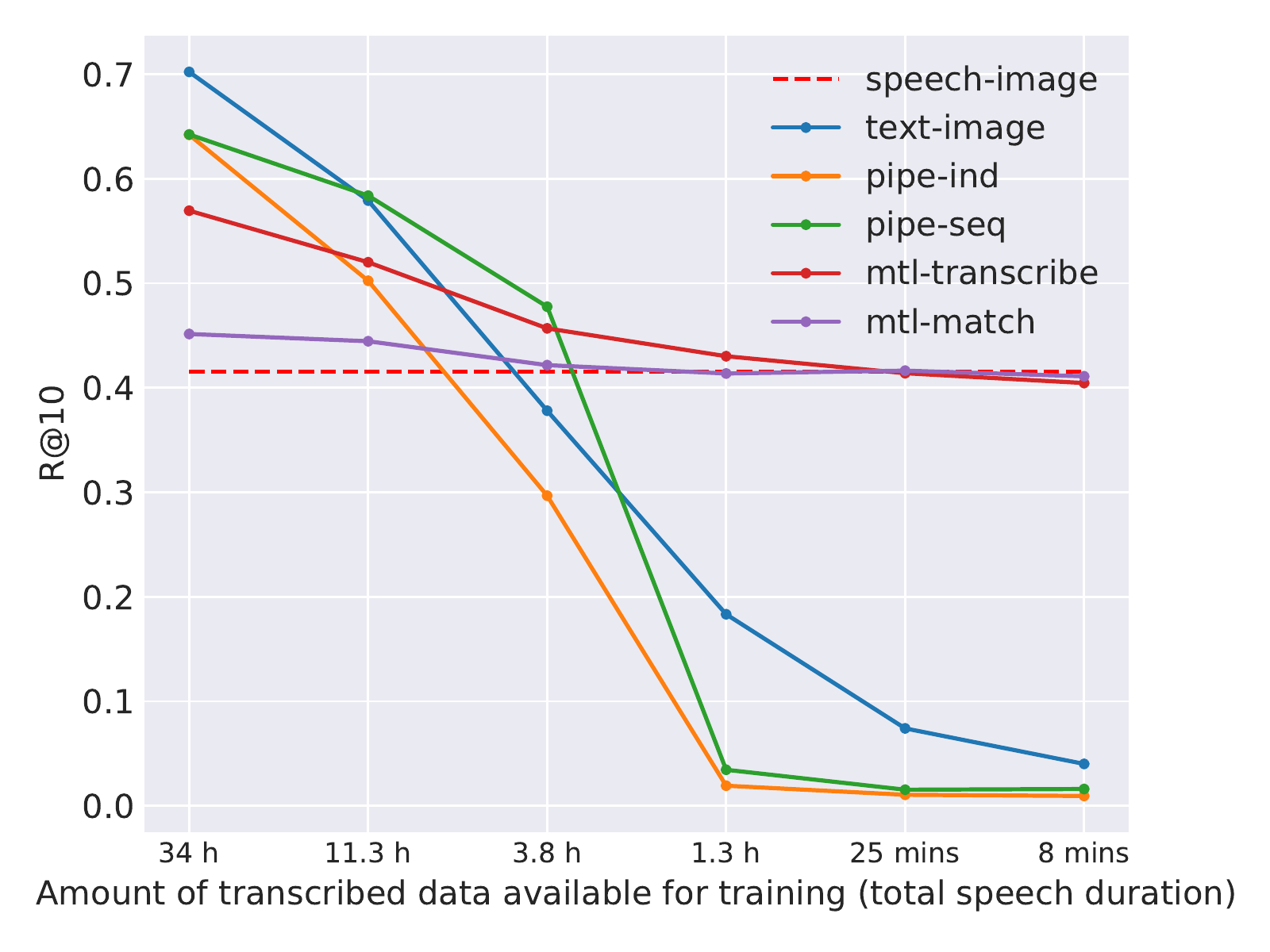}
    \end{subfigure}
    \begin{subfigure}[t]{0.5\linewidth}
        \includegraphics[width=\linewidth]{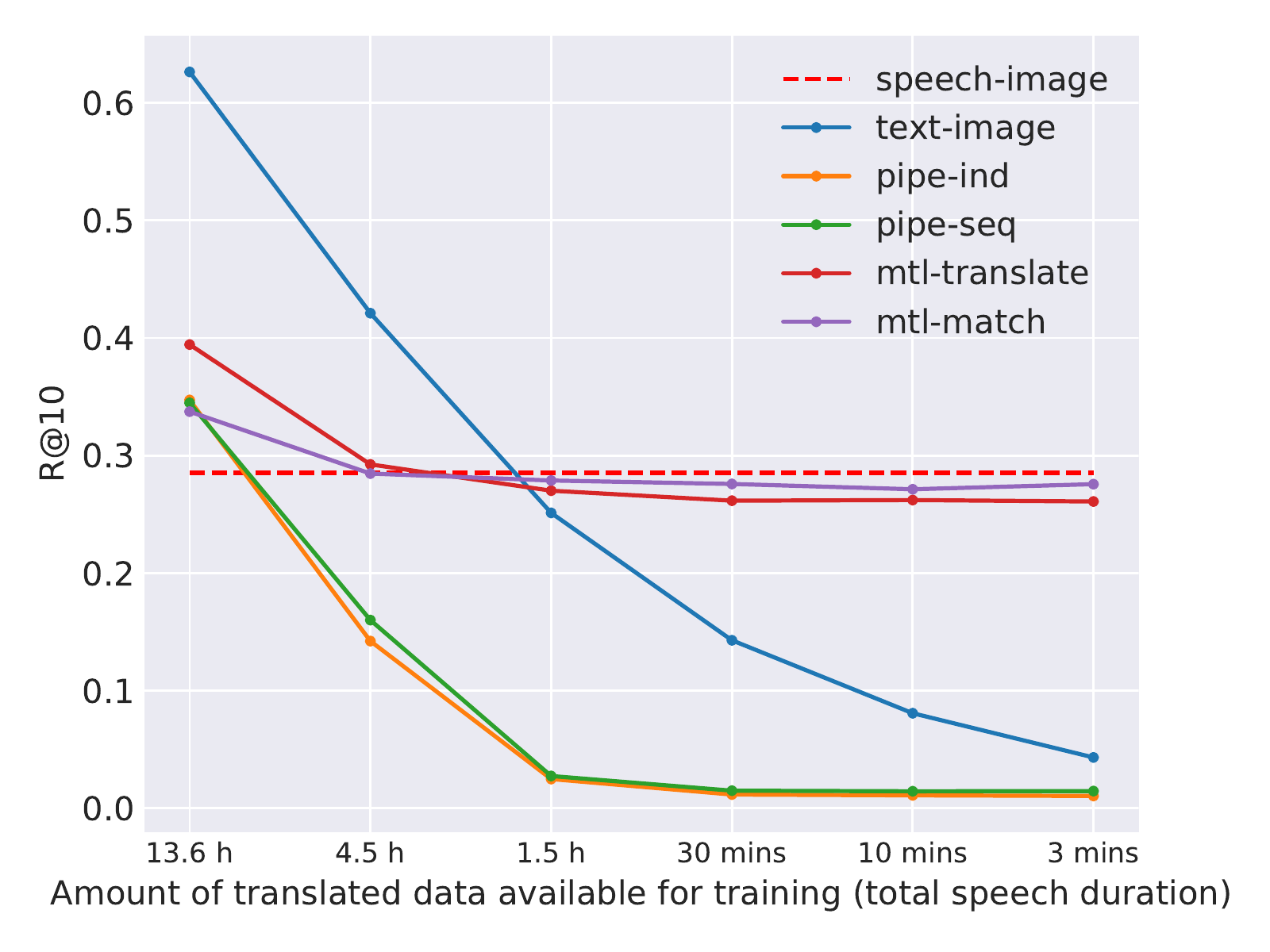}
    \end{subfigure}
    \caption{Results (R@10) of the different models on the validation set, when
        trained with decreasing amounts of English transcriptions (left) or
        Japanese translations (right). The total amount of speech available is
        kept identical, only the amount of translated data changes.}
    \label{fig:downsampling_text}
\end{figure*}

We can see in Figure \ref{fig:downsampling_text} (left) that, as the amount of
transcribed data decreases, the score of the \emph{text-image} and pipeline
models progressively goes toward $0\%$ of R@10. Between $11.3$ and $3.8$ hours
of transcribed data, the \emph{text-image} and \emph{pipe-ind} models fall
below the performance of the \emph{speech-image} baseline. The \emph{pipe-seq}
is more robust and its performance stays higher (the best actually) until the
amount of data goes below $3.8$ hours of transcribed speech. After that the
R@10 falls abruptly. This is likely the effect of the sequential training
procedure allowing the \emph{pipe-seq} to use all the speech available to train
the text-image module (by transcribing it with the \gls{ASR} module). Below
$3.8$ hours of speech, the quality of the transcriptions given by the \gls{ASR}
system deteriorates to the point that it is not usable anymore by the downstream
component.

The two multitask approaches, on the other hand, progressively converge toward
the speech-only baseline. The \emph{mtl-transcribe} approach is overall giving
better results than the \emph{mtl-match} approach but fails to give a
significant advantage over other sytems. It is only after the performance of
the \emph{pipe-seq} system abruptly decreases (from $1.3$ hours of transcribed
speech and below) that the \emph{mtl-transcribe} system can surpass this one,
at which point it is already performing very close to the \emph{speech-image}
baseline.

Figure \ref{fig:downsampling_text} (right) reports on the same set of experiments
with Japanese translations. In this case too, the
\emph{text-image}, \emph{pipe-ind} and \emph{pipe-seq} models go toward $0\%$
of R@10 as the amount of translated data decreases, while the
\emph{mtl-translate} and \emph{mtl-match} systems converge toward the
\emph{speech-image} baseline. It seems though that $4.5$ hours of translated
data is not enough to see an improvement over the \emph{speech-image} baseline
with any of the models.

In this case, the \emph{pipe-seq} model does not have a significant
advantage over the \emph{pipe-ind} model, likely due to the difficulty of the
translation task. The same reason probably explains why the
\emph{mtl-transcribe} strategy is performing the best on the full dataset (as
reported in Section \ref{sec:res_trl}). However, while the pipeline architectures
never surpass the \emph{mtl-translate} model in the experiments reported, it may be the case with more data.

For completeness, Appendix \ref{sec:app_asr} reports the performance of
on the \gls{ASR} and \gls{SLT} tasks themselves, for decreasing amounts of
textual data.

\section{Conclusion}

In this paper, we investigated the use of textual supervision in
visually-grounded models of spoken language understanding. We found
that the improvements reported in \citet{chrupala_symbolic_2019} and
\citet{pasad_contributions_2019} are a low estimate of what can be
achieved when textual labels are available. Among the different
approaches we explored, the more traditional pipeline approach,
trained sequentially, is particularly effective and hard to beat with
end-to-end systems. This indicates that text is a very powerful
intermediate representation.  End-to-end approaches tend to perform better
only when the amount of textual data is limited.

We have also shown that written translations are a viable alternative to
transcriptions (especially for unwritten languages), though more data might be
useful to compensate for the harder task.

\subsection{Limitations and future work}

We ran our experiments on Flickr8K dataset, which is a read speech dataset. We
are thus likely underestimating the advantages of end-to-end approaches over
pipeline approaches, in that they can use information present in speech (such
as prosody) but not in text. Running experiments on a dataset with more
natural and conversational speech could show end-to-end systems in a better
light.

On the other end, we restricted ourselves to training the \gls{ASR} and
\gls{NLU} components of the pipeline systems independently. Recent techniques
such as Gumbel-Softmax \citep{jang_categorical_2017} or the
straight-through estimator \citep{bengio_estimating_2013} could be applied to
train/finetune this model in an end-to-end fashion while still enforcing a
symbolic intermediate representation similar to text.

In the same vein, it would be interesting to explore more generally whether and
how an inductive bias could be incorporated in the architecture to encourage
the model to discover such kind of symbolic representation naturally.

\section*{Acknowledgements}
Bertrand Higy was supported by a NWO/E-Science Center grant
number~027.018.G03.

\bibliographystyle{acl_natbib}
\bibliography{refs}

\begin{comment}
% hacky way to have avoid tables on a separate page
\begin{table*}[!t]
    \centering
    \begin{tabular}{ l|cccccc }
        \hline
        Amount of transcribed data & 34 h & 11.3 h & 3.8 h & 1.3 h & 25 mins & 8 mins \\
        \hline
        \Acrlong{WER} & 0.154 & 0.238 & 0.397 & 0.801 & 1.03 & 0.977 \\
        \hline
    \end{tabular}
    \caption{Performance (\gls{WER}) of the \gls{ASR} component on the
    validation set, when trained with decreasing amount of transcribed data.}
    \label{tab:res_asr}
\end{table*}
\begin{table*}[!t]
    \centering
    \begin{tabular}{ l|cccccc }
        \hline
        Amount of translated data & 13.6 h & 4.5 h & 1.5 h & 30 mins & 10 mins & 3 mins \\
        \hline
        BLEU score & 0.256 & 0.153 & 0.073 & 0.065 & 0.040 & 0.021 \\
        \hline
    \end{tabular}
    \caption{Performance (BLEU score) of the \gls{SLT} component on the
    validation set, when trained with decreasing amount of translated data.}
    \label{tab:res_slt}
\end{table*}
\end{comment}
\begin{table*}
    \centering
    \begin{tabular}{ l|p{5cm} }
        \noalign{\hrule height 1pt}
        Component & Number of layers \\
        \noalign{\hrule height 1pt}
        text decoder & \{1, \textbf{2}, 3\} \\
        \hline
        speech encoder of the \gls{ASR} module & \{3, 4, 5, 6\} \\
        \hline
        speech encoder of the \gls{SLT} module & \{4, \textbf{5}, 6\} \\
        \hline
        speech encoder of the \emph{mtl-transcribe} model & \{(2, 2, 2),
        (3, 1, 1), (3, 2, 2), (4, 0, 0), (4, 0, 1), (4, 1, 0), \textbf{(4, 1,
        1)}, (4, 1, 2), (4, 2, 1), (4, 2, 2), (5, 0, 0), (5, 1, 1)\} \\
        \hline
        speech encoder of the \emph{mtl-translate} model & \{(3, 1, 1),
        \textbf{(4, 1, 1)}, (4, 2, 2), (5, 1, 1), (5, 2, 2), (6, 1, 1)\} \\
        \noalign{\hrule height 1pt}
    \end{tabular}
    \caption{List of values we experimented with for the number of \gls{GRU}
    layers in the different components. The best configuration is indicated in
    bold face.}
    \label{tab:num_layers}
\end{table*}
\begin{table*}
    \centering
    \begin{tabular}{ l|p{5cm} }
        \noalign{\hrule height 1pt}
        Component & Dimension of the hidden state \\
        \noalign{\hrule height 1pt}
        \emph{speech-image} model & \{256, 512, 768, \textbf{1024}\} \\
        \hline
        \gls{ASR} module & \{512, \textbf{768}, 1024\} \\
        \hline
        \emph{mtl-transcribe} model & \{512, \textbf{768}, 1024\} \\
        \noalign{\hrule height 1pt}
    \end{tabular}
    \caption{List of values we experimented with for the number of \gls{GRU} layers in the different components. The best configuration is indicated in bold face.}
    \label{tab:num_units}
\end{table*}
\newpage
%\mbox{}
%\newpage
%\end{comment}

%\clearpage
\appendix
\section{Appendices}
\label{sec:appendix}

\subsection{Choice of hyperparameters}
\label{sec:app_hyperparameters}

The hyperparameters related to the optimization procedure and the architecture
of the \emph{speech-image} model where chosen based on
\citet{merkx_language_2019}. While the architecture of the other components 
largely follows this baseline, the number of \gls{GRU} layers and the
dimension of their hidden state were manually tuned to optimize accuracy. An
exception to this is the \emph{mtl-match} model for which the number of layer
of the speech and text encoders is taken from \citet{chrupala_symbolic_2019}.
Optimization is done based on single runs.

\subsubsection{Number of \gls{GRU} layers}

Table \ref{tab:num_layers} reports the values we experimented with for the
number of \gls{GRU} layers in the text encoder, the speech encoder of the
\gls{ASR} module and the speech encoder of the \gls{SLT} module. For the
\emph{mtl-transcribe} and \emph{mtl-translate} systems, we report the triplet
corresponding to the number of the layers in the shared encoder, the encoder
dedicated to the speech-image task and the encoder dedicated to the
transcription/translation task.

\begin{comment}
\begin{table*}
    \centering
    \begin{tabular}{ l|p{5cm} }
        \noalign{\hrule height 1pt}
        Component & Number of layers \\
        \noalign{\hrule height 1pt}
        text decoder & \{1, \textbf{2}, 3\} \\
        \hline
        speech encoder of the \gls{ASR} module & \{3, 4, 5, 6\} \\
        \hline
        speech encoder of the \gls{SLT} module & \{4, \textbf{5}, 6\} \\
        \hline
        speech encoder of the \emph{mtl-transcribe} model & \{(2, 2, 2),
        (3, 1, 1), (3, 2, 2), (4, 0, 0), (4, 0, 1), (4, 1, 0), \textbf{(4, 1,
        1)}, (4, 1, 2), (4, 2, 1), (4, 2, 2), (5, 0, 0), (5, 1, 1)\} \\
        \hline
        speech encoder of the \emph{mtl-translate} model & \{(3, 1, 1),
        \textbf{(4, 1, 1)}, (4, 2, 2), (5, 1, 1), (5, 2, 2), (6, 1, 1)\} \\
        \noalign{\hrule height 1pt}
    \end{tabular}
    \caption{List of values we experimented with for the number of \gls{GRU}
    layers in the different components. The best configuration is indicated in
    bold face.}
    \label{tab:num_layers}
\end{table*}
\end{comment}

\subsubsection{Dimension of the hidden state of the \gls{GRU} layers}

Table \ref{tab:num_units} reports the values we experimented with for the
dimension of the hidden state of the \gls{GRU} layers in the
\emph{speech-image} model, the \gls{ASR} component and the
\emph{mtl-transcribe} model. The best value for the \emph{speech-image} model
was reused for the \emph{text-image} and \emph{mtl-match} models, as well as
the text-image component of the pipeline models. The best value for the
\gls{ASR} module and the \emph{mtl-transcribe} model was reused for the
\gls{SLT} module and the \emph{mtl-translate} model.

\begin{comment}
\begin{table*}
    \centering
    \begin{tabular}{ l|p{5cm} }
        \noalign{\hrule height 1pt}
        Component & Dimension of the hidden state \\
        \noalign{\hrule height 1pt}
        \emph{speech-image} model & \{256, 512, 768, \textbf{1024}\} \\
        \hline
        \gls{ASR} module & \{512, \textbf{768}, 1024\} \\
        \hline
        \emph{mtl-transcribe} model & \{512, \textbf{768}, 1024\} \\
        \noalign{\hrule height 1pt}
    \end{tabular}
    \caption{List of values we experimented with for the number of \gls{GRU} layers in the different components. The best configuration is indicated in bold face.}
    \label{tab:num_units}
\end{table*}
\end{comment}

\subsection{Complete set of results}
\label{sec:app_res}

We report here on the performance of the models presented in section
\ref{sec:results} (Tables \ref{tab:res_baselines}, \ref{tab:res_core} and
\ref{tab:res_test}) with additional metrics, namely R@1, R@5 and Medr.  Tables
\ref{tab:res_baselines_full} and \ref{tab:res_core_full} report the performance
on the validation set of the two reference and the four core models
respectively. Table \ref{tab:res_test_full} reports the performance on the test
set of the best pipeline and multitask models. Performance appears consistent
accross metrics.

\begin{table*}[t]
    \centering
    \begin{tabular}{ lccccccccc }
        \hline
        Model & \multicolumn{3}{c}{Transcriptions (full)} & \multicolumn{3}{c}{Transcriptions (reduced)} & \multicolumn{3}{c}{Translations} \\
              & R@1 & R@5 & Medr & R@1 & R@5 & Medr & R@1 & R@5 & Medr \\
        \hline
        \emph{speech-image} & 0.105 & 0.299 & 16.2 & 0.059 & 0.188 & 38.3 & 0.059 & 0.192 & 36.0 \\
        \emph{text-image} & 0.258 & 0.566 & 4.0 & 0.228 & 0.508 & 5.0 & 0.209 & 0.492 & 6.0 \\
        \hline
    \end{tabular}
    \caption{Validation set performance (R@1, R@5 and Medr) of our two
    reference models when trained either with English transcriptions, a reduced
    version of the English dataset, or Japanese translations.}
    \label{tab:res_baselines_full}
\end{table*}

\begin{table*}[t]
    \centering
    \begin{tabular*}{\textwidth}{ lccccccccc }
        \hline
    Model & \multicolumn{3}{c}{\makecell{Transcriptions\\(full)}} & \multicolumn{3}{c}{\makecell{Transcriptions\\(reduced)}} & \multicolumn{3}{c}{Translations} \\
              & R@1 & R@5 & Medr & R@1 & R@5 & Medr & R@1 & R@5 & Medr \\
        \hline
        \emph{pipe-ind} & 0.232 & 0.514 & 5.0 & 0.187 & 0.452 & 7.0 & 0.088 & 0.248 & 27.3 \\
        \emph{pipe-seq} & 0.224 & 0.509 & 5.2 & 0.190 & 0.459 & 7.0 & 0.082 & 0.242 & 27.0 \\
        \emph{mtl-transcribe}/\emph{mtl-translate} & 0.177 & 0.431 & 7.8 & 0.133 & 0.352 & 12.0 & 0.091 & 0.285 & 19.3 \\
        \emph{mtl-match} & 0.115 & 0.321 & 13.0 & 0.079 & 0.244 & 24.3 & 0.071 & 0.232 & 26.3 \\

        \hline
    \end{tabular*}
    \caption{Validation set performance (R@1, R@5 and Medr) of our four core
    models, when trained either with English transcriptions, a reduced version
    of the English dataset, or Japanese translations.}
    \label{tab:res_core_full}
\end{table*}

\begin{table*}[t]
    \centering
    \begin{tabular}{ lcccccc }
        \hline
        Model & \multicolumn{3}{c}{Transcriptions} & \multicolumn{3}{c}{Translations} \\
              & R@1 & R@5 & Medr & R@1 & R@5 & Medr \\
        \hline
        \emph{pipe-seq} & 0.218 & 0.499 & 6.0 & 0.079 & 0.248 & 26.5 \\
        \emph{mtl-transcribe}/\emph{mtl-translate} & 0.174 & 0.425 & 8.0 & 0.099 & 0.279 & 19.0 \\
        \hline
    \end{tabular}
    \caption{Test set performance (R@1, R@5 and Medr) of the best pipeline
    (\emph{pipe-seq}) and the best multitask
    (\emph{mtl-transcribe}/\emph{mtl-translate}) models, when trained with all
    English transcriptions or all Japanese translations.}
    \label{tab:res_test_full}
\end{table*}

\subsection{Performance of the \gls{ASR} and \gls{SLT} systems}
\label{sec:app_asr}

Tables \ref{tab:res_asr} and \ref{tab:res_slt} respectively report the
performance of the \gls{ASR} and \gls{SLT} modules on their own tasks,
when trained on decreasing amount of textual data. Evaluation is performed on
the validation set with a beam of width 10.

%\begin{comment}
\begin{table*}[!t]
    \centering
    \begin{tabular}{ l|cccccc }
        \hline
        Amount of transcribed data & 34 h & 11.3 h & 3.8 h & 1.3 h & 25 mins & 8 mins \\
        \hline
        \Acrlong{WER} & 0.154 & 0.238 & 0.397 & 0.801 & 1.034 & 0.977 \\
        \hline
    \end{tabular}
    \caption{Performance (\gls{WER}) of the \gls{ASR} component on the
    validation set, when trained with decreasing amount of transcribed data.}
    \label{tab:res_asr}
\end{table*}
\begin{table*}[!t]
    \centering
    \begin{tabular}{ l|cccccc }
        \hline
        Amount of translated data & 13.6 h & 4.5 h & 1.5 h & 30 mins & 10 mins & 3 mins \\
        \hline
        BLEU score & 0.256 & 0.153 & 0.073 & 0.065 & 0.040 & 0.021 \\
        \hline
    \end{tabular}
    \caption{Performance (BLEU score) of the \gls{SLT} component on the
    validation set, when trained with decreasing amount of translated data.}
    \label{tab:res_slt}
\end{table*}
%\end{comment}

\end{document}